\documentclass[11pt]{article}
\usepackage{emnlp2016}
\usepackage{times}
\usepackage{latexsym}

\usepackage{CJKutf8}
\emnlpfinalcopy 

\usepackage{examples}
\usepackage{balance}
\usepackage{cgloss4e}
\usepackage{verbatim}
\usepackage{cprotect}
\usepackage{listings}
\usepackage{url}
\usepackage{amsmath}
\usepackage{stackrel}
\usepackage[pdftex]{graphicx}
\usepackage{dblfloatfix}
\usepackage{graphics}
\usepackage{array}
\newcolumntype{x}[1]{>{\centering\arraybackslash\hspace{0pt}}p{#1}}
\newcolumntype{y}[1]{<{\hspace{-.15cm}}p{#1}}

\usepackage[usenames,dvipsnames,svgnames,table]{xcolor}

\usepackage{hyperref}

\usepackage{CJKutf8}
\usepackage{ucs}
\usepackage{CJKulem} 
\usepackage[utf8]{inputenc}
\newenvironment{SChinese}{%
  \CJKfamily{gbsn}%
  \CJKnospace
  \CJKtilde}{}
\newenvironment{Japanese}{%
  \CJKfamily{min}%
  \CJKnospace
  \CJKtilde}{}

\usepackage{color}

\usepackage{amssymb,amsmath,epsfig}
\usepackage{mathpartir}
\usepackage{proof}
\usepackage{amsthm}
\usepackage{xspace}
\usepackage{algorithm}
\usepackage[noend]{algpseudocode}
\algrenewcommand{\algorithmicindent}{1em}

\usepackage{enumitem}

\newcommand{\notes}[1]{}

 \theoremstyle{definition}
 
\theoremstyle{plain}

\newcommand{\ith}[1]{\ensuremath{i^{{th}}}}

\newcount\permx
\newcount\permy
\def\permdot#1#2{
\permx=#1 \advance\permx by-1
\permy=#2 \advance\permy by-1
\psframe[fillcolor=black, fillstyle=solid]
(\permx,\permy)(#1, #2)
}

\newcommand{\boxnum}[1]{{\setlength{\fboxsep}{1pt}\raisebox{1pt}{\hspace{1pt}\fbox{\tiny #1}\hspace{1pt}}}}
\newcommand{\ind}[1]{\ensuremath{_{\kern-0.5pt\boxnum{#1}}}}

\def\namecite{\newcite}

\newcommand{\smallnt}[1]{\ensuremath{_{\mbox{\tiny PP}}}\xspace}

\newcommand{\pseudocode}{Algorithm}
\floatname{algorithm}{\pseudocode}

\iffalse

\else

\fi

\newcommand{\ter}{{\sc Ter}\xspace}
\newcommand{\bleu}{{\sc Bleu}\xspace}
\newcommand{\ribes}{{\sc Ribes}\xspace}

\usepackage{multirow}
\usepackage{tikz}
\usetikzlibrary{positioning}
\usepackage{tikz-qtree}
\usetikzlibrary{arrows}

\title{Temporal Attention Model for Neural Machine Translation}

\author{Baskaran Sankaran \quad Haitao Mi \quad Yaser Al-Onaizan \quad Abe Ittycheriah \\
IBM T.J.~Watson Research Center \\
1101 Kitchawan Rd, Yorktown Heights, NY 10598 \\
{\tt \{bsankara, hmi, onaizan, abei\}@us.ibm.com}
}
\date{}

\begin{document}
\maketitle
\begin{CJK*}{UTF8}{}

\begin{abstract}
Attention-based Neural Machine Translation (NMT) models suffer from 
\textit{attention deficiency} issues as has been observed in recent research. 
We propose a novel mechanism to address some of these limitations and improve the NMT attention. 
Specifically, our approach memorizes the alignments temporally (within each sentence)
and modulates the attention with the accumulated temporal memory, as the decoder 
generates the candidate translation. 
We compare our approach against the baseline NMT model and 
two other related approaches that address this issue either explicitly or implicitly. 
Large-scale experiments on two language pairs 
show that our approach achieves better and robust gains 
over the baseline and related NMT approaches. Our model further outperforms strong SMT baselines in some settings even without using ensembles.

\end{abstract}

\section{Introduction}
\label{sec:intro}
Neural machine translation (NMT) is gaining significant interest in recent years~\cite{kalchbrenner+blunsom:2013,sutskever+:2014,bahdanau+:2014,jean+:2015,luong+:2015b}. They have also been successful in reaching performance comparable to the traditional statistical machine translation (SMT) models and has attained state of the art performances against the SMT for certain language pairs~\cite{jean+:2015,luong+:2015b}, albeit with ensembling. \\

Most of the recent models are driven by some form of attention originally proposed by~\namecite{bahdanau+:2014}. The attention mechanism enables the model to dynamically compose source representations for each timestep during decoding, instead of a single, static representation. Specifically the mechanism allows the model to accord varying attention to different parts of source sentence while generating successive target words. 

While the attention mechanism remembers the source words over fairly long distances, the attention in each step only looks at the previous hidden state
and the source annotations. The model does not directly encode information as to which source words/ fragments have been focused in the earlier timesteps. This deficiency manifests in two distinct but related issues, where the attention i) focuses on same source word/ fragment beyond what is needed, leading to repeated words in the translation and ii) fails to focus on some part of the input and thus missing it completely in the translation leading to \textit{lack} of adequacy.

\begin{table*}[htb]
{\footnotesize
\begin{center}
\begin{tabular}{p{1.5cm} p{13.5cm}}
\hline
\textit{Source:} & die Teilnehmer der Proteste , die am Donnerstag um 6:30 AM morgens vor dem McDonald 's in der 40th Street und in der Madison Avenue begannen , forderten , dass \textbf{die Kassierer und K\"{o}che von Fast - Food - Restaurants} einen Mindestlohn von 15 US-Dollar die Stunde erhalten , was mehr als einer Verdoppelung des jetzigen Mindestlohns entspricht . \\
\textit{Reference:} & Participants of the protest that began at 6.30 a.m. on Thursday near the McDonald's on 40th street and Madison Avenue demanded that cashiers and cooks of the fast food chain be paid at least 15 dollars/hour, i.e. more than double their present wages. \\
\textit{Candidate:} & The protests that began on Thursday at 06:30 before the McDonald 's at \underline{McDonald 's at} \underline{McDonald 's} on 40th Street and Madison Avenue demanded that a minimum wage of 15 dollars would receive \underline{a minimum wage of 15 dollars} per hour , equivalent to doubling the current minimum wage . \vspace{0.5mm} \\
\hline

\textit{Source:} & Special attention \textbf{is being paid} to the Tokyo \textbf{gubernatorial election} because it is perceived as a litmus test for the upcoming House of Councillors election , particularly in the metropolitan areas where \textbf{nonpartisan voters predominate} . \\
\textit{Reference:} & \begin{CJK}{UTF8}{min}都 知事 選 の 結果 は , とくに 参院 選 に 向け 都市 部 に 多い 無党派 層 の 動向 を 占う もの と して 注目 さ れる 。\end{CJK} \\
\textit{Candidate:} & \begin{CJK}{UTF8}{min}特に 首都 圏 で は , \underline{特に 首都 圏 で は ,} 参院 選 の 試金石 と して 注目 さ れて いる 。\end{CJK} \vspace{0.5mm} \\
\hline
\end{tabular}
\end{center}
\caption{\label{tbl:attn_limitation} Examples for phrase repetitions and inadequacy issues in NMT shown for German-English and English-Japanese settings. The repeated phrases in the translations are shown by \underline{underlined} text, while the missing source content are \textbf{boldfaced}. Notice that the repetitions need not be consecutive, making the issue even more critical.}
}
\end{table*}

Table~\ref{tbl:attn_limitation} shows illustrative examples of these issues for 
two distinct language pairs. It is interesting to note that sometimes the system repeats even a fairly long phrase as we seen in the Japanese example. 
In reality, the NMT model can repeat the fragments several times; we have recorded as high as 8 repetitions in certain cases. Additionally, the system completely misses some salient parts of the sentence; for example it frequently fails to translate the verbs in Japanese. The problem is exacerbated for longer sentences or for language pairs with complex reordering. These issues has attracted considerable interest in the recent past~\cite[\textit{inter alia}]{luong+:2015b,xu+:2015}. 

Unlike the earlier works, we take a different approach in this paper and directly save the attention at each timestep. We then use this temporal memory along with that of the source annotations to modulate the attention to produce the response at the current timestep. The use of temporally aggregated alignments can be seen as a memory-network inspired extension to the NMT attention. 

The model naturally allows one-to-many and many-to-many alignments to be captured if supported by the data (effectively modeling high fertility words), but otherwise overriding them if the weights are small (thus preventing repetitions). In contrast to the alternative approaches, our model is simpler as it avoids additional parameters.
Large-scale experiments on two language pairs show that our approach improves the translation quality over the baseline NMT system significantly, and also surpasses two other related approaches.

We first give an outline of the NMT model with attention~\cite{bahdanau+:2014} in~\autoref{sec:nmt}. 
We then explain our augmented attention model employing temporal alignments (\autoref{sec:hist_aligns}). We then turn to discuss two alternative models for improving attention i) coverage embedding (\autoref{sec:cov}) inspired by the traditional SMT and ii) local attention models (\autoref{sec:local}). We go over the related work in~\autoref{sec:related} and then present our experiments and results in~\autoref{sec:exps} before concluding.

\section{Neural Machine Translation}
\label{sec:nmt}
\begin{figure*}[!t]
\centering
\includegraphics[width=1\textwidth]{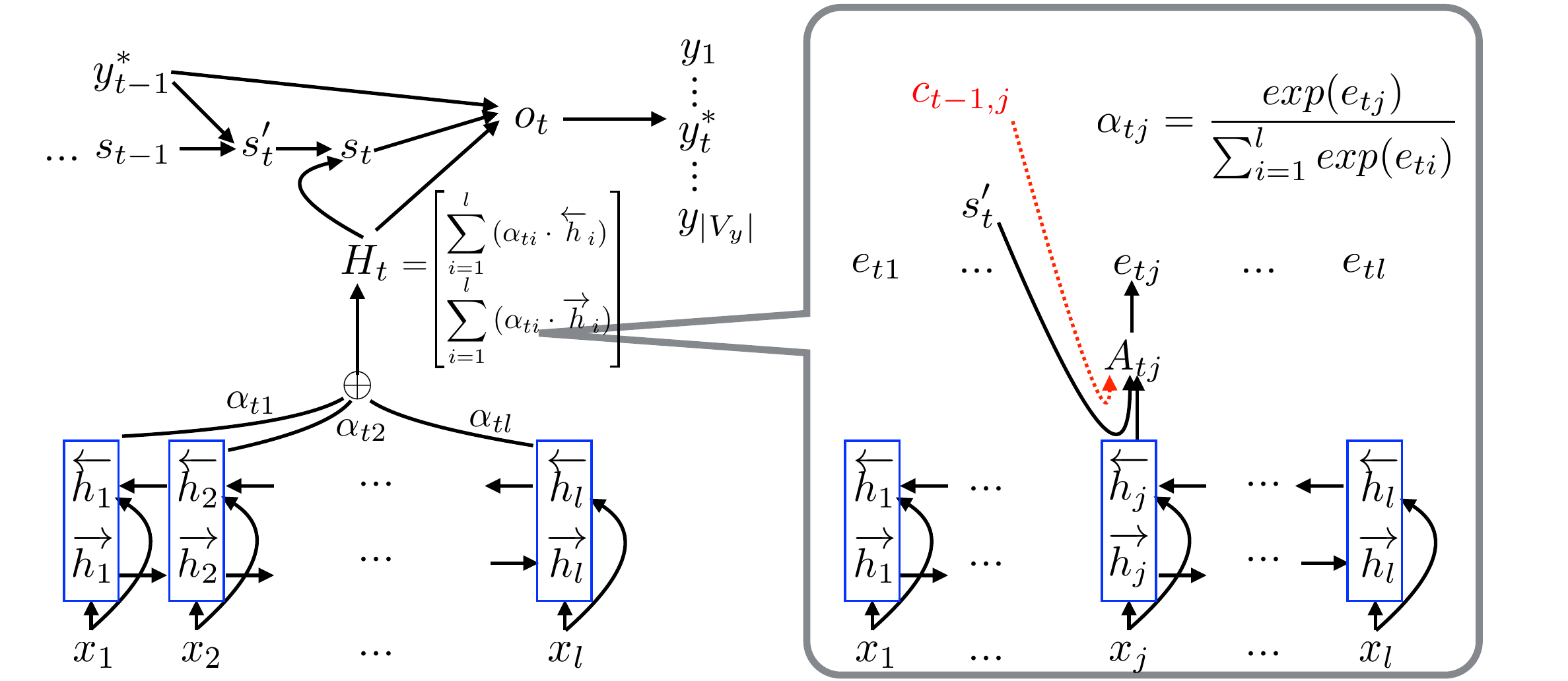}
\caption{Attention-based NMT architecture with the attention model illustrated within the box on the right. Unlike~\protect\namecite{bahdanau+:2014} our architecture employs a conditional GRU layer that has two GRU units surrounding an attention mechanism.\protect\footnotemark \, See the accompanying text for description. The dotted (red) connection is used in the coverage-embedding model (Sec.~\protect\ref{sec:cov}).}
\label{fig:att}
\end{figure*}

As shown in Figure~\ref{fig:att}, 
attention-based NMT~\cite{bahdanau+:2014} is an encoder-decoder network.  
The encoder employs a bi-directional RNN to 
encode the source sentence ${\bf{x}}=({x_1, ... , x_l})$ 
into a sequence of hidden states ${\bf{h}}=({h_1, ..., h_l})$, where $l$ is the length of the source sentence.
Each $h_i$ is a concatenation of a left-to-right $\overrightarrow{h_i}$
and a right-to-left $\overleftarrow{h_i}$ RNN:
\[
h_{i} = 
\begin{bmatrix}
\overleftarrow{h}_i \\ 
\overrightarrow{h}_i \\
\end{bmatrix}
=
\begin{bmatrix}
\overleftarrow{f}(x_i, \overleftarrow{h}_{i+1}) \\
\overrightarrow{f}(x_i, \overrightarrow{h}_{i-1}) \\
\end{bmatrix}
\]
where $\overleftarrow{f}$ and $\overrightarrow{f}$ 
are two gated recurrent units (GRU) introduced by~\namecite{cho+:2014_gru}.

Given the encoded ${\bf h}$, the decoder predicts the target translation
by maximizing the conditional log-probability of the 
correct translation ${\bf y^*} = (y^*_1, ... y^*_m)$, where 
$m$ is the length of the target. At each time $t$, 
the probability of each word $y_t$ from a target vocabulary $V_y$ is:
\begin{equation}
\label{eq:py}
p(y_t|{\bf h}, y^*_{t-1}..y^*_1) = g(s_t, y^*_{t-1}, H_{t}),
\end{equation}
where $g$ is 
a two layer feed-forward network ($o_t$ being an intermediate state) 
over the embedding of the previous target word ($y^*_{t-1}$),  
the decoder hidden state ($s_t$), and the weighted sum of encoder states ${\bf h}$ ($H_{t}$). A single feedforward layer then projects $o_t$ to the target vocabulary and applies softmax to predict the probability distribution over the output vocabulary.

We compute $s_t$ with a two layer GRU as:
\begin{equation}
s'_t = u(s_{t-1}, y^*_{t-1}).
\end{equation}
\begin{equation}
s_t = q(s'_{t}, H_{t})
\end{equation}
\noindent where $s'_t$ is an intermediate state.
The two GRU units $u$ and $q$ together with the attention constitute the conditional GRU layer. And $H_{t}$ is computed as:
\begin{equation}
H_t = 
\begin{bmatrix}
\sum_{i=1}^{l}{(\alpha_{t,i} \cdot \overleftarrow{h}_i)} \\
\sum_{i=1}^{l}{(\alpha_{t,i} \cdot \overrightarrow{h}_i)} \\
\end{bmatrix},
\end{equation}

\footnotetext{Same as the decoder GRU introduced in session-2 of the dl4mt-tutorial: https://github.com/nyu-dl/dl4mt-tutorial/tree/master/session2.}

The attention model (in the right box) is a two layer feed-forward network $r$, with $A_{t,j}$ being an intermediate state and another layer converting it into a real number $e_{t,j}$. The alignment weights $\alpha$, are computed from the two layer feed-forward network $r$ as:
\begin{equation}
\alpha_{t,i} = \frac{\exp\{r(s'_{t}, h_{i})\}}{\sum_{j=1}^{l}{\exp\{r(s'_{t}, h_{j})\}}}
\end{equation}

$\alpha_{t,j}$ are actually the soft alignment probabilities, denoting the probability of aligning the target word at timestep $t$ to source position $j$.

\section{Temporal Attention Model}
\label{sec:hist_aligns}

\begin{figure}[ht]
\centering
\includegraphics[width=0.4\textwidth]{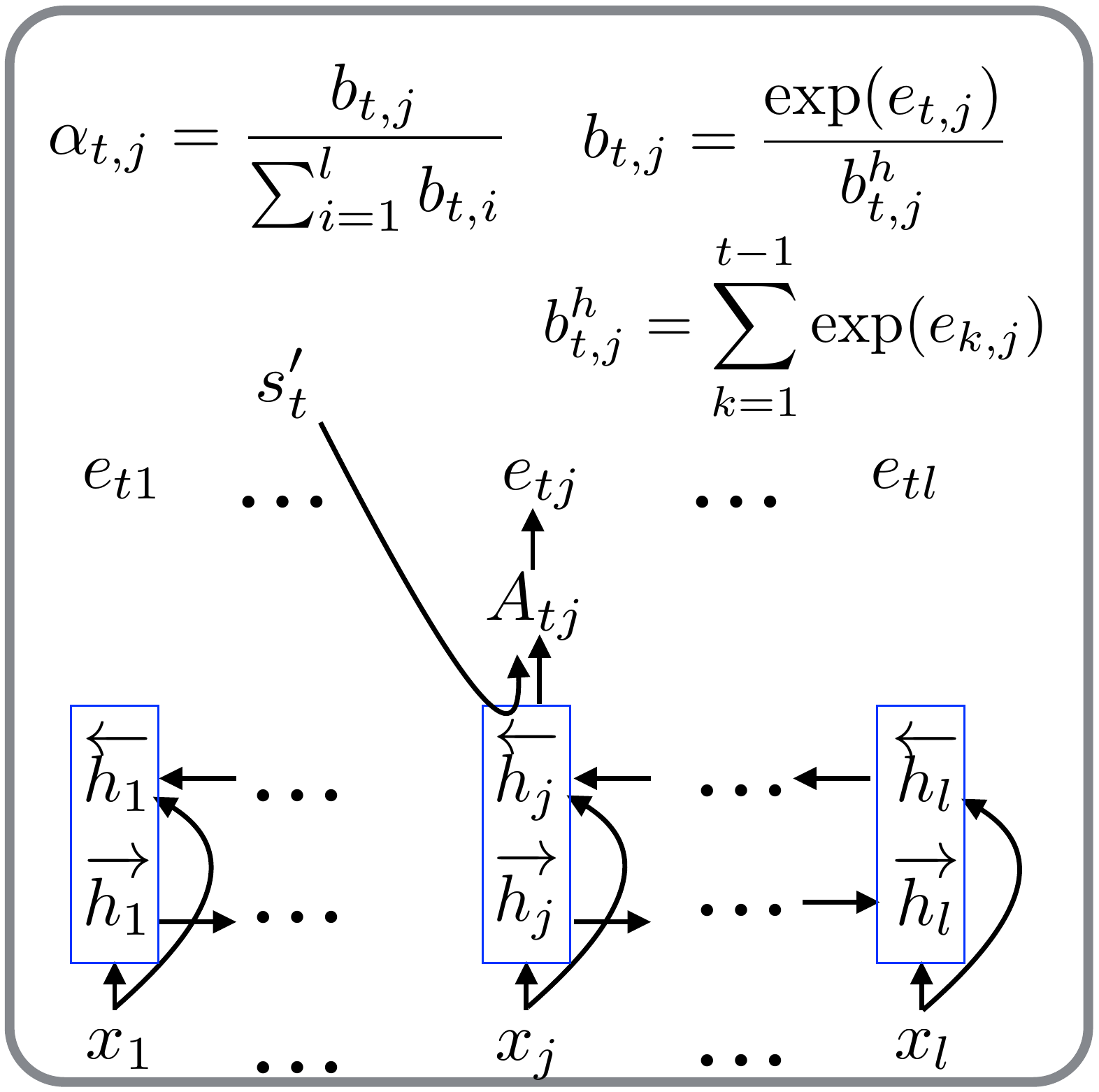}
\caption{Temporal Attention model:
$b_{t,j}$ is the alignment value prior to normalization. We modulate the attention
by looking at historical alignments $b^h_{t}$ 
followed by a normalization over all the source positions.
Intuitively, we take into account the entire attention matrix 
instead of just one dimension.
}
\label{fig:hist}
\vspace{-0.25cm}
\end{figure}

In this section, we propose a novel mechanism to memorize the alignments temporally from the previous timesteps and then use those to modulate the attention in the subsequent timesteps. Intuitively we would like the attention to be \textit{aware} of what it focussed on in the previous steps. The vanilla attention model expects this to be captured by longer short-term memory in the decoder hidden state. However this only encodes indirect information about the alignments in previous timestep(s). In our approach, we propose to directly use the historical alignment information to better modulate the attention.

Figure~\ref{fig:hist} illustrates the temporally modulated alignment $b_{t,j}$ at decoder timestep $t$ and source postion $j$. The $e_{t,j}$ are modulated based on the historical alignments $b^h_{t,j}$:
\begin{equation}
b^h_{t,j} = \sum_{k=1}^{t-1} \exp(e_{k,j})
\end{equation}

\noindent where $b^h_{t,j}$ denotes the aggregated alignments until (but not including) the current timestep $t$. The modulated alignment $b_{t,j}$ then takes the form:
\begin{equation}
b_{t,j} = \frac{\exp(e_{t,j})}{b^h_{t,j}}
\end{equation}

\noindent However at timestep $t = 1$, we set the modulated alignment to be $b_{1,j} = \exp(e_{1,j})$ as there is no previous history.
We then compute the attention weights $\alpha_{t,j}$ by normalizing along the input sequence length:
\begin{equation}
\alpha_{t,j} = \frac{b_{t,j}}{\sum_{i=1}^{l} b_{t,i}}
\end{equation}

In this way, we explicitly force the model to remember \textit{all} its previous decisions, which we believe, will be particularly useful in language pairs with complex reordering\footnote{We also tried adding an extra connection instead from temporal alignments and learn its parameters, but that performed poorly than directly modulating the attention.}.

It would be interesting to see how the model performs if we limit the temporal attention to the last $n$ timesteps, instead of the entire history. Thus the model can \textit{forget} older decisions, which arguably might not be relevant going forward (we do not address this question here, but leave it for future experimentation).

\subsection{A Perspective from Memory Networks}
Neural Memory Networks~\cite{graves+:2014,weston+:2014,sukhbaatar+:2015} have gained substantial interest in past two years. The key idea in MNs is to augment the neural networks to effectively interface with memory so that the network can selectively choose to perform read or write operations as required. While the memory content are encoded by the hidden states and weights, the power of such models manifest in their ability to \textit{memorize}/ \textit{recall} facts to/ from the memory potentially over long term and/ or large volume (of encoded facts).

The NMT attention can be considered as a simple example for MNs, where the memory stores the source annotation at the sentence level. Our approach could be seen from the perspective of memory networks (MNs) in that we further extend the attention to \textit{memorize} its decisions temporally during decoding.

Compared to the typical MNs, our model differs in terms of what, where and how we \textit{memorize}. In our work, the memory is \textit{volatile} as we only use the aggregated alignments within that sentence; it is not reused when the decoder moves to the next one. Secondly, we do not store the alignments externally or by gating it; instead we directly use it to modulate the alignments. 
Finally, while the conventional memory networks would have one or more layers of memory, whose parameters are trained as part of the model~\cite{sukhbaatar+:2015}, 
our model memorizes the alignment vectors (and not the model parameters) in the attention.

\section{Alternative Models for Attention}
In this section we discuss two alternative approaches that address deficiency in the attention model either explicitly or implicitly.
\subsection{Coverage Embedding Model}
\label{sec:cov}

\begin{figure}[!t]
\centering
\includegraphics[width=0.4\textwidth]{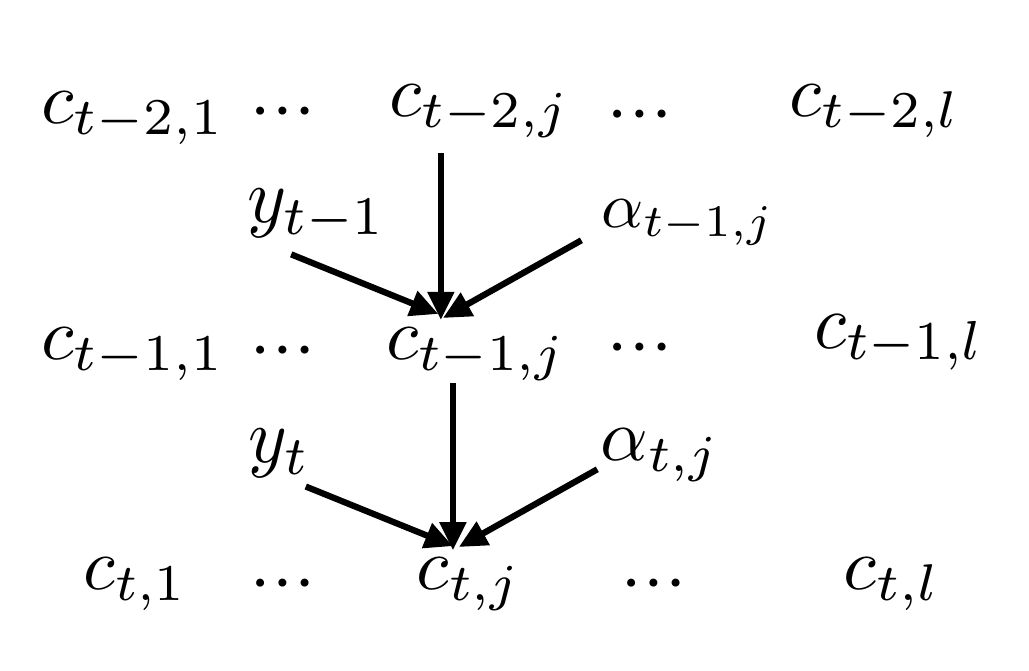}
\vspace{-0.5cm}
\caption{The coverage embedding model with a GRU at time step $t-1$ and $t$. 
$c_{0,1}$ to $c_{0,l}$ are initialized with the word coverage embedding matrix.}
\label{fig:cov}
\vspace{-0.4cm}
\end{figure}



Traditional statistical machine translation (e.g. \cite{koehn:2004}) decoders usually employ a source side ``coverage vector'' to keep track of the source words that are already translated. 
At the beginning the coverage vector is initialized to be all zeros, which means no word has been translated.
Once the word at source position $j$ is translated, its index in the coverage vector is set to 1; and the decoder won't translate the word again in that sentence. 

The coverage embedding model~\cite{mi+:2016} seeks to adapt the notion of ``coverage vector'' for NMT. Under this, each source word $x_i$ is assigned a coverage embedding,\footnote{Following~\namecite{mi+:2016}, we set $c_{t,j}$ dimension to 100.} which intuitively encodes its fertility\footnote{Fertility of a word is its yield in target language measured by the \# of words} to be one of the three possible cases: one-to-one, one-to-many or one-to-zero.

Figure~\ref{fig:cov} illustrates the coverage embedding model.
During training, at timestep $0$, the coverage embeddings of the input sequence are initialized by table look-up from the coverage embedding matrix, yielding $c_{0,1}, c_{0,2}, ... c_{0,l}$. At time step $t$, the model uses the coverage embedding $c_{t-1,j}$ in the attention model (shown by the dotted line in Figure~\ref{fig:att}). 
The $c_{t,j}$ is subsequently updated by the attention weights $\alpha_{t,j}$, predicted target $y_{t}$ and embedding from the previous timestep $c_{t-1,j}$ (see Figure~\ref{fig:cov}) with a GRU as:
\begin{equation*}
\begin{split}
& z_{t,j} = \sigma(W^{zy}y_{t} + W^{z\alpha}\alpha_{t,j} + U^{z}c_{t-1,j}) \\
& r_{t,j} = \sigma(W^{ry}y_{t} + W^{r\alpha}\alpha_{t,j} + U^{r}c_{t-1,j}) \\
& \tilde{c}_{t,j} = \tanh(W y_{t} + W^{\alpha} \alpha_{t,j} + r_{t,j} \circ U c_{t-1,j})\\
& c_{t,j} = z_{t,j} \circ c_{t-1, j} + (1-z_{t,j}) \circ \tilde{c}_{t,j}, \\
\end{split}
\end{equation*}
where, $z_t$ is the update gate, $r_t$ is the reset gate, 
$\tilde{c}_t$ is the new memory content, and $c_t$ is the final memory.
The matrix $W^{zy}$, $W^{z\alpha}$, $U^{z}$, $W^{ry}$, $W^{r\alpha}$, $U^{r}$,
$W^{y}$, $W^\alpha$ and $U$ are shared across different position $j$.
$\circ$ is a pointwise operation. 



\subsection{Local Attention Model}
\label{sec:local}

The local attention~\cite{luong+:2015b} was originally motivated as an alternative to the vanilla \textit{global} attention model~\cite{bahdanau+:2014}. Unlike the global attention, which attends to \textit{all} source words at each timestep, the local attention selectively limits the focus to a \textit{local} window of source sentence at each step. The window approach thus avoids the probability mass being allocated to the positions outside the window and so the resulting attention can be expected to be sharper than the global attention. We compare our temporal attention to the local attention model as they both attempt to modulate the attention, albeit in distinct ways.

The two variants of the local attention differ in whether the local window at each timestep is chosen based on i) monotone assumption (local-$m$) between source and target or ii) a parameterized and predictive model (local-$p$) that predicts the source alignment position for the given timestep. Interestingly the published work employs the local attention only for En-De translation; for the reverse De-En direction it only shows the results for global attention (Table 3 in~\namecite{luong+:2015b}). It is not clear if the local attention was helpful at all in the latter setting. 
We only consider local-$m$ attention for our experiments in this paper. 


\section{Related Work}
\label{sec:related}
Recently \namecite{xu+:2015} proposed a doubly stochastic attention model for image caption generation, which is very close to our work in terms of its motivation. The decoder generates a description as the attention model attends to a specific location in the image. While the attention at every step sums to $1$ ($\sum_i \alpha_{ti} = 1$), the same is not true for across all the timesteps for a fixed position $i$. Thus, the decoder can completely ignore some parts of the image (equivalently sentences in the case of NMT). To address this, they add a penalty term to the cost, which \textit{softly} encourages the attention to focus on every part of the image. In contrast, our model enforces this explicitly and provides a stronger prior than the soft penalty. Our experiments with the penalty term did not perform better than the baseline.

The notion of coverage vector in the phrase-based SMT has inspired similar approaches for NMT~\cite{tu+:2016,mi+:2016}. The former employs a GRU to model the coverage vector, which are initialized with a uniform distribution. This model is very similar to the coverage-embedding model (described in~\autoref{sec:cov}) that we compare against in our experiments. One main difference is that the latter model initializes each source word from a specific coverage embedding matrix. Secondly~\namecite{tu+:2016} add an accumulate operation and a fertility function to simulate the one-to-many alignments scenario, whereas~\cite{mi+:2016} add fertility information directly to coverage embeddings, as each source word has its own embedding.

Alternately~\namecite{feng+:2016} proposes an additional recurrent layer that has explicit connections from previous attention RNN hidden state, previous target label and current attention generated context. While this model retains the previous attention information, it is only through the attention generated context (in contrast, we memorize the attention weights across each timestep). They also condition the decoder during training to explicitly model the distortion and fertility through additional terms to the cost. However the major drawback is that their approach was tested only on a small dataset ($0.5M$ sentences) and hence it is not clear if the improvements would hold when applied to large datasets.

\namecite{cohn+:2016} take a different approach and augment the attention model with well-known features in traditional SMT, including positional bias, Markov conditioning, fertility and bilingual symmetry feature. The features are added as extra components to the pre-normalized attention where the features are computed dynamically from the parallel sentences and the weights are learned through backprop. The last bilingual symmetry feature is captured by an additional term in the loss function. 
They only experiment their model on simulated low-resource setting and so it is unclear if the gains carry over to the large data NMT.

In the context of continuous speech recognition (SR),~\namecite{chorowski+:2014} proposed a soft alignment model based on NMT attention as part of a hybrid end-to-end SR system. The NN component was used for generating phonemes from the raw speech signals, wherein the attention model included the relative position information and a penalty term to constrain the attention in a roughly monotone path. While the one-to-one, monotone relationship between input and output serves well for speech recognition, this will not be applicable for the NMT.


\section{Experiments}
\label{sec:exps}

\subsection{Datasets}
We run our experiments on: Chinese-English (Zh-En) and German-English (De-En).

\noindent \textbf{Zh-En}: The training corpus is from the DARPA BOLT Chinese-English task consisting of about $5M$ sentences and includes a mix of newswire, broadcast news, webblog from various sources. Our dev set is the concatenation of several tuning sets, 
having 4491 sentences. 
Three test-sets are NIST MT06 (1664), MT08 news (691), and MT08 web (666). 

\noindent \textbf{De-En}: The training corpus consists of about $4M$ sentence pairs, which is a sub-set of WMT14.
The development and test sets are WMT-13 and WMT-14 testsets (3000 sentences each) respectively. 



\subsection{NMT Experimental Setting}
For the NMT systems, we use the vocabulary sizes of $300k$ and $60k$ for the Zh-En and De-En respectively. We used the embedding dimension of 620 for De-En and 300 for the other two language pairs\footnote{Reducing the embedding dimensions allows us to speed up training especially given the larger output vocabulary sizes.}. We fix the RNN GRU layers to be of 1000 cells each. 
We use AdaDelta~\cite{zeiler:2012} or SGD\footnote{Most of the times, we found them to be comparable in performance across different languages. And we do not present a detailed comparison here as it is beyond the scope of this paper.} for optimization and update the model parameters with a mini-batch size of 80.

For the large vocabulary Zh-En setting, we use a modified vocabulary manipulation method~\cite{jean+:2015} during training and restrict the output vocabulary for each mini-batch with a \textit{candidate list}. In generating the candidate list, we used the top $2k$ most frequent target words along with top-$k$\footnote{We fixed $k=10$} candidates from the respective word-to-word translation table learned using `fast\_align'~\cite{dyer+:2013}. We also augment the candidate list with the tree-to-string phrases, where for many-to-many phrases\footnote{We fixed the maximum source side length to be $4$.}, we assumed each word in source phrase to be aligned to every other word in the target side of the rule. Following~\namecite{jean+:2015}, we also added the words from the target sides of the training data to the candidate list only during training, to ensure that the translations are reachable.

During translation, we dump the alignments (from the attention mechanism for each timestep for each sentence), and use these alignments to replace the UNK tokens either with potential targets (obtained from the word translation table) or with the source word (if no target was found).

\begin{table*}[htpb]
\centering
\tabcolsep=0.2cm
\begin{tabular}{l c | c c | c c | c c}
\multicolumn{2}{l|}{\multirow{3}{*}{System}} & \multicolumn{2}{c|}{\multirow{2}{*}{MT06}}  & \multicolumn{4}{c}{MT08} \\ \cline{5-8} 
                       &              & &                       & \multicolumn{2}{c|}{News}  & \multicolumn{2}{c}{Web} \\ \cline{3-8} 
     &                           & BP   & \bleu(TB)     & BP   & \bleu(TB)     & BP & \bleu(TB) \\ 
\hline \hline
\multicolumn{2}{l|}{SMT}    	      & 0.95 & 34.93 \, (\textit{9.45})  & 0.94 & \textit{31.12} (\textit{12.90}) & 0.90 & 23.45 (17.72) \\ 
\hline
\multicolumn{2}{l|}{LVNMT}         & 0.96 & 34.53 (12.25) & 0.93 & 28.86 (17.40) & 0.97 & 26.78 (17.57) \\ 
\multicolumn{2}{l|}{\, + Temporal Attention}& 0.92 & \textbf{36.34 \, (9.95)} & 0.88 & \textbf{31.49 (13.93)} & 0.94 & \textbf{27.29} (\textbf{16.04}) \\ 
\hline
\multicolumn{2}{l|}{\namecite{mi+:2016}}    & 0.91 & 35.34 (10.78) & 0.88 & 29.85 (15.38) & 0.96 & \textbf{27.35} (16.67) \\ 
\hline\hline
\multicolumn{2}{l|}{\namecite{tu+:2016}\footnotemark} & N/A & 32.47\;\; (N/A)  & \multicolumn{4}{c}{ N/A\;\;\; 25.23 (N/A)} \\
\hline
\end{tabular}
\caption{Zh-En: the brevity penalty (BP), \bleu$\uparrow$ and TB~$\downarrow$ scores (within brackets). TB (the lower the score the better the translation) measures (\ter-\bleu)/2 and is agnostic to longer translations which will not be penalized by \bleu. NMT results are on a large vocabulary ($300k$) and with UNK replacement. The best scores for all NMT systems in each test set are in \textbf{boldface} and the better SMT scores are \textit{italisized}. LVNMT stands for NMT model trained with large vocab manipulation. \label{tbl:zhen}}
\vspace{-0.3cm}
\end{table*}


Considering the morphological richness of German, we follow the approach by~\namecite{sennrich+:2015} and segment the German and English sides into subwords by using byte-pair encoding~\cite{gage:1994}. We segment the two corpora separately with distinct vocabularies.\footnote{Byte-pair encoding learned on the combined German and English corpus did not perform as well.} Thus the De-En NMT system does not have any UNK tokens. 

\footnotetext{\protect\namecite{tu+:2016} is not comparable to others as they use 1.25 million training set, small vocabulary size $30k$, tune on NIST 2002 test set, and tune directly to maximize \bleu.}

\subsection{SMT Baselines}
Our traditional SMT baseline is a hybrid syntax-based tree-to-string model~\cite{zhao+yaser:2008}.
For Zh-En task, we first parse the Chinese side with Berkeley parser, then align the bilingual sentences with GIZA++ and finally extract Hiero/ tree-to-string rules for translation.

The language models for are trained on the English side of the bitexts along with a large  monolingual corpora (around $10b$ words from Gigaword (LDC2011T07) and Google News). 
We tune our SMT system with PRO~\cite{hopkins+may:2011} to minimize (\ter - \bleu)/2 (TB) on the dev set.

\subsection{Evaluation}
We use two ways to evaluate our models. 
First we present translation results for automatic metrics \bleu and TB. We use the dev set to choose the best model, and then report test set scores for this model.

Then, we also assess the alignment quality of the competing NMT approaches for Zh-En. 
Given the source and target sentence pairs in hand-aligned alignment datasets,
we forced decode with the NMT models to produce the reference translations while recording the alignments from the attention layer. We then compute the Precision and Recall by comparing the machine-generated and gold alignments.

\begin{table}[ht]
\centering
\begin{tabular}{l c c}
System  & \bleu$\uparrow$ & TB~$\downarrow$ \\
\hline \hline
SMT            & 26.47 & 13.85 \\
\hline
NMT            & 24.91 & 15.60 \\
\, + Ensemble & 26.99 & 13.54 \\
\hline
\namecite{mi+:2016}     & 25.74 & 14.94 \\ \hline
Local Attn~\cite{luong+:2015b}  & 26.42 & 14.59 \\
\, + Ensemble & 28.20 & 12.95 \\ \hline
Temporal Attention & \textbf{26.66} & \textbf{14.26} \\
\, + Ensemble & \textit{28.32} & \textit{12.72} \\ \hline
\end{tabular}
\caption{De-En: \bleu and TB~$\downarrow$ scores on the WMT-2014 test set. NMT systems are trained with $60k$ byte-pair encoded vocabulary. Best single systems are \textbf{boldfaced} and while the best ensemble scores (from 4 models) are \textit{italisized} \label{tbl:deen}}
\end{table}

\begin{table*}[htb]
{\footnotesize
\begin{center}
\begin{tabular}{p{1.5cm} p{13.5cm}}
\hline
\textit{Temporal Attn:} & The protests that began on Thursday at 06:30 in front of the McDonald's in 40th Street and Madison Avenue demanded that the casks and cooks of fast-food restaurants receive a minimum wage of USD 15 per hour , which is more than doubling the current minimum wage . \\
\hline

\textit{Temporal Attn:} & \begin{CJK}{UTF8}{min}特に , 参院 選 で の 選挙 で は , 特に 有権者 の 優位 性 の ある 都市 圏 で は , 特に 注目 さ れて いる 。\end{CJK}\\ 
\hline
\end{tabular}
\end{center}
\caption{\label{tbl:attn_limitation_resolved} Output from our Temporal Attention Model for examples in Table~\ref{tbl:attn_limitation} showing better translations}
}
\end{table*}

\subsubsection{Translation Evaluation}

The main results are summarized in Tables~\ref{tbl:zhen} and~\ref{tbl:deen}, and we begin with Zh-En results. As has been noted by others, the baseline NMT systems generally perform poorly than their SMT counterparts (MT08-Web testset is an exception).
 
The coverage embedding model (labelled~\namecite{mi+:2016}) improves both \bleu and TB scores across by up to $1$ and $2$ points respectively. Our temporal attention model improve this further by a maximum of $1.4$ TB point. The \bleu scores show similar gains of about $1.5$ points; interestingly the \bleu scores of our proposed model are even better than the traditional SMT (MT06 and MT08-Web) and is comparable for the MT08-news testset.

The last row of the table lists the \bleu scores from~\namecite{tu+:2016}. While our \bleu scores are not directly comparable because of the differences in the training set, we do note here that our scores are 3 points better than theirs on the MT06 and MT08 testsets in absolute terms. Further, our SMT system offers a very strong baseline for two reasons: i) it is a Hiero-style system, which performs much better than a corresponding phrase-based model for Zh-En, ii) it employs a robust language model trained not just with the target side of the bitext but with large monolingual English data as we noted earlier.

For the De-En setting, we compare our temporal attention model with the coverage-emdedding as well as local attention. The coverage embedding model improves the TB score by $0.65$. The local attention model performs better than the coverage embedding improving the \bleu by $0.7$. Our proposed historical alignment model improves it further by about $0.35$ TB. Finally, we tested ensemble models for baseline, local-attention and temporal attention NMT systems, which yield $1.7$ to $2$ \bleu points improvement over the corresponding single systems. Overall, the local attention and historical alignments ensemble models outperform SMT by $\sim{}2$ \bleu. 
We present the results for En-Jp setting (not discussed due to lack of space) that shows our model considerably improving over baseline NMT.


Table~\ref{tbl:attn_limitation_resolved} lists the output from our proposed model for examples in Table~\ref{tbl:attn_limitation}, showing better translations without repetitive phrases. We also calculated the no. of repetitive phrases for De-En (not tabulated for need of space). We observed our model to produce $40\%$ fewer repetitions compared to local attention. Average length of repetitions reduced from $7.27$ for local attention to $3.47$ for temporal attention.

\begin{table}
\centering
\begin{tabular}{l c c}
System  & \bleu$\uparrow$ & \ribes$\uparrow$ \\
\hline \hline
SMT            & \textit{32.06} & 0.6894 \\
LVNMT            & 27.54 & 0.7103 \\ 
\, + Temporal Attention & \textbf{28.70} & \textbf{0.7232} \\ 
\hline
\end{tabular}
\caption{En-Jp: Trained on $16M$ sentences bitext. NMT systems use $100k$ LV model with the candidate translations from model-1 table along with top $2k$ most freq target words. Shows \bleu and \ribes~\protect\cite{isozaki+:2010} scores.\label{tbl:enjp}}
\vspace{-0.3cm}
\end{table}

\begin{table}
\centering
\begin{tabular}{l c c c}
System  & Prec. & Rec. & F1 \\
\hline \hline
MaxEnt & 74.86 & 77.10 & 75.96 \\
LVNMT  & 47.88 & 41.06 & 44.21 \\
\, + Temporal Attention & 50.33 & \textbf{43.57} & \textbf{46.71} \\
\namecite{mi+:2016} & \textbf{51.11} & 41.42 & 45.76 \\
\hline
\end{tabular}
\caption{Zh-En: Alignment $F1$ scores for different models on 447 hand-aligned sentences. \label{tbl:align_zhen}}

\vspace{-0.3cm}
\end{table}

%
%


\subsubsection{Alignment Evaluation}
Now we turn to evaluate the alignments generated by our temporal attention model; we do this in the Zh-En setting using 447 hand-aligned sentence pairs. We compare against the coverage embedding as well as the baseline LVNMT. Additionally we also use a maximum entropy (MaxEnt) aligner trained on $67k$ hand-aligned data (Table~\ref{tbl:align_zhen}).

Both coverage embedding and historical alignments approaches improve the $F1$ score by $1.55$ and $2.5$ respectively over the baseline NMT model. The temporal attention model gets highest recall and we hypothesize that this is because the historical alignments are better in capturing many-to-many alignments. 
However all the NMT models are way behind the maximum-entropy aligner.

\section{Conclusion and Future Work}
\label{sec:future_extn}
We propose an extension to the NMT attention model, where the model memorizes the alignments temporally and aggregates them in the memory allowing the attention model to remember its decisions over past timesteps. 
We showed better performance on two language pairs over the NMT baseline and also against two alternative NMT models employing coverage embedding and local attention.

It would be interesting to see if our temporal alignments model could be extended further to memorize the alignments for the phrase-pairs across individual sentence pairs. 
Secondly, the local attention model is complementary to our approach and hence it is natural to combine the two.

\section*{Acknowledgment}
This work is supported by DARPA HR0011-12-C-0015 (BOLT), and
The views and findings in this paper are those of the authors 
and are not endorsed by the DARPA.

The views, opinions, and/or findings contained in this article/presentation 
are those of the author/presenter and should not be interpreted as representing 
the official views or policies, either expressed or implied, of the DARPA.

\balance
\bibliographystyle{emnlp2016}
\bibliography{temp_align}

\end{CJK*}

\end{document}